\documentclass{article}

\usepackage{booktabs,siunitx,multirow}
\usepackage{enumitem}
\usepackage[preprint, nonatbib]{neurips_2024}

\usepackage[utf8]{inputenc} 
\usepackage[T1]{fontenc}    
\usepackage{hyperref}       
\usepackage{url}            
\usepackage{booktabs}       
\usepackage{amsfonts}       
\usepackage{nicefrac}       
\usepackage{microtype}      
\usepackage{xcolor}         
\definecolor{deepgreen}{rgb}{0.0, 0.5, 0.0} 

\usepackage{graphicx}

\title{Prompt-Consistency Image Generation (PCIG): A Unified Framework Integrating LLMs, Knowledge Graphs, and Controllable Diffusion Models}

\author{%
  Yichen Sun, Zhixuan Chu$^{*}$, Zhan Qin$^{*}$, Kui Ren\\
   \text{Zhejiang University}  \\
  \texttt{\{yichensun,zhixuanchu,qinzhan,kuiren\}@zju.edu.cn}
   \\
}

\begin{document}

\maketitle
\def\thefootnote{*}\footnotetext{Corresponding author.}

\begin{abstract}
  The rapid advancement of Text-to-Image(T2I) generative models has enabled the synthesis of high-quality images guided by textual descriptions. Despite this significant progress, these models are often susceptible in generating contents that contradict the input text, which poses a challenge to their reliability and practical deployment. To address this problem, we introduce a novel diffusion-based framework to significantly enhance the alignment of generated images with their corresponding descriptions, addressing the inconsistency between visual output and textual input. Our framework is built upon a comprehensive analysis of inconsistency phenomena, categorizing them based on their manifestation in the image. Leveraging a state-of-the-art large language module, we first extract objects and construct a knowledge graph to predict the locations of these objects in potentially generated images. We then integrate a state-of-the-art controllable image generation model with a visual text generation module to generate an image that is consistent with the original prompt, guided by the predicted object locations. Through extensive experiments on an advanced multimodal hallucination benchmark, we demonstrate the efficacy of our approach in accurately generating the images without the inconsistency with the original prompt. The code can be accessed via \href{https://github.com/TruthAI-Lab/PCIG}{https://github.com/TruthAI-Lab/PCIG}.
\end{abstract}
\section{Introduction}
The rapid advancement of Text-to-Image (T2I) generative models has revolutionized the field of computer vision, enabling the synthesis of high-quality images guided by textual descriptions. These models, such as DALL-E \cite{DALL-E3,ramesh2021zero}, Stable Diffusion \cite{podell2023sdxl}, and GLIDE \cite{nichol2021glide}, have shown remarkable progress in generating visually appealing and semantically relevant images. However, despite their impressive performance, these models often generate contents that contradict the input text, posing significant challenges to their reliability and practical deployment.
\begin{figure*}[t]
    \centering
    \includegraphics[width=0.7\textwidth]{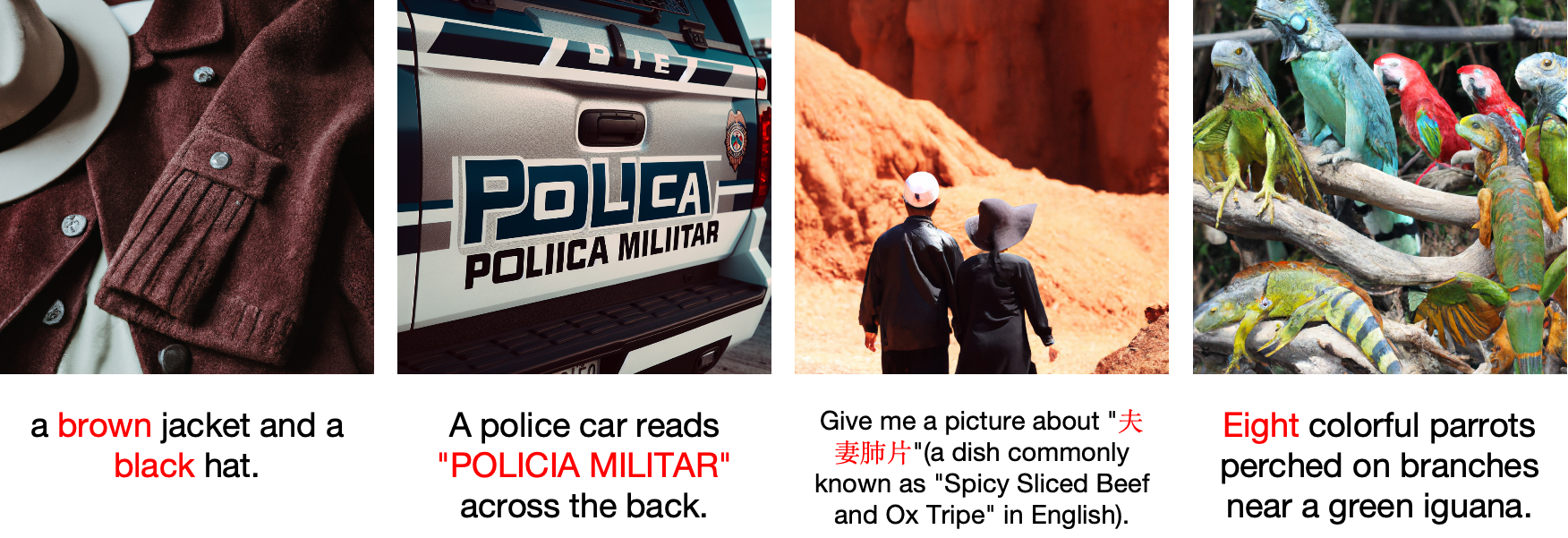}
    
    \caption{Selected samples generated by DALL-E 3. Each image represents one specific hallucination type. The inconsistency part for each image is highlighted in red.}
    \vspace{-3mm}
    \label{hall}
    
\end{figure*}
Inconsistencies between the visual output and textual input can manifest in various forms, such as mismatched object attributes (First image in Figure \ref{hall}), inaccurate object placement or count (Fourth image in Figure \ref{hall}), illegible or incorrect text within the image (Second image in Figure \ref{hall}), and the inability to accurately depict real-world entities (Third image in Figure \ref{hall}). These inconsistencies, also known as hallucinations, can severely impact the usefulness and trustworthiness of the generated images, especially in domains where accuracy is crucial, such as medical imaging \cite{kazerouni2023diffusion,liu2024survey}, autonomous vehicles \cite{liu2024ddm}, and criminal investigation \cite{nowroozi2021survey}.

Existing methods have attempted to address these challenges by improving the alignment between the input text and the generated image. Attention-based approaches \cite{ho2020denoising,song2020denoising} have been proposed to better capture the relationships between words and visual features, while adversarial training techniques \cite{frolov2021adversarial} have been employed to enhance the realism and consistency of the generated images. However, these methods often struggle with complex scenes and fail to address the specific types of inconsistencies mentioned earlier. 

To tackle these limitations, we introduce Prompt-Consistency Image Generation (PCIG), a novel diffusion-based framework that significantly enhances the alignment of generated images with their corresponding descriptions. PCIG addresses three key aspects of consistency: (1) general objects (GO), ensuring accurate depiction of object attributes and placement; (2) text within the image (TEXT), generating legible and correct text; and (3) objects that refer to proper nouns existing in the real world (PN), which cannot be directly generated by the model.

Our framework leverages state-of-the-art techniques in natural language processing and computer vision. We first employ large language models (LLMs) \cite{gpt4} to extract objects from the input prompt and construct a knowledge graph to predict the locations of these objects in the generated image. LLMs, such as GPT-3 \cite{brown2020language} and BERT \cite{devlin2018bert}, have shown remarkable capabilities in understanding and generating human language. By integrating LLMs into our framework, we enable a deeper understanding of the prompt and its relationships, guiding the subsequent image generation process.

Next, we utilize a controllable diffusion model \cite{li2023gligen,wang2024instancediffusion,zhou2024migc} to generate an image consistent with the original prompt, guided by the predicted object locations. Controllable diffusion models allow for more fine-grained control over the image generation process by incorporating additional constraints or conditions. For general objects (GO), the model focuses on accurate attribute depiction and spatial arrangement. To handle text within the image (TEXT), we incorporate a visual text generation module \cite{tuo2023anytext,ma2023glyphdraw,yang2024glyphcontrol} that specializes in rendering legible and semantically correct text. Recent advances in visual text generation have shown promising results in producing realistic and readable text in images. Finally, for objects referring to proper nouns (PN), we propose a novel approach that searches for representative images of the entities and seamlessly integrates them into the generated image.

Through extensive experiments on an advanced multimodal hallucination benchmark \cite{chen2024unified}, we demonstrate the efficacy of PCIG in generating images that align with the original prompt, significantly reducing inconsistencies across all three key aspects. Our unified framework achieves state-of-the-art performance, outperforming existing T2I models in terms of object hallucination accuracy, textual hallucination accuracy, and factual hallucination accuracy.

\paragraph{Contributions.}
(1) We introduce PCIG, a novel framework that integrates LLMs, knowledge graphs, and controllable diffusion models to generate prompt-consistent images.
    (2) We propose a comprehensive approach to address three key aspects of consistency: general objects, text within the image, and objects referring to proper nouns.
    (3) We conduct extensive experiments on a multimodal hallucination benchmark, demonstrating the superiority of PCIG over existing T2I models in terms of consistency and accuracy.
   (4) We provide insights into the effectiveness of integrating LLMs and knowledge graphs for prompt understanding and object localization in the image generation process.
\section{Related Work}
\subsection{Text-to-image Diffusion Models}
Denoising Diffusion Probabilistic Model \cite{ho2020denoising,song2020denoising} and its subsequent studies \cite{ho2022classifier,ramesh2022hierarchical,saharia2022photorealistic,rombach2022high,nichol2021glide,ramesh2021zero} have showcased impressive capabilities in generating high-quality images guided by textual prompts. These models employ iterative denoising steps starting from a random noise map to learn the process of text-to-image generation. Latent Diffusion Model (LDM) \cite{rombach2022high} takes advantage of iterative denoising steps in a latent space, aiming to enhance text-to-image alignment and reduce training complexity while generating high-quality images from textual descriptions. Stable Diffusion and SDXL \cite{podell2023sdxl} are applications of the Latent Diffusion method in text-to-image generation but trained with additional data and a powerful CLIP \cite{radford2021learning} text encoder. DALL-E 2 \cite{ramesh2021zero} and DALL-E 3 \cite{DALL-E3}, state-of-the-art text-to-image generation model developed by OpenAI, achieve photorealistic T2I generation using diffusion-based models.
\subsection{Controllable Image Generation}
As text description cannot precisely control the position of generated instances, Some controllable text-to-image generation methods \cite{gafni2022make,li2023gligen,avrahami2023spatext,bar2023multidiffusion,zhou2024migc,wang2024instancediffusion,zhang2023adding,xie2023boxdiff} introduce spatial conditioning controls to guide the image generation process. They extend the pre-trained T2I model \cite{rombach2022high} to integrate layout information into the generation and achieve control of instances’ position. GLIGEN \cite{li2023gligen}, MIGC \cite{zhou2024migc}, and InstanceDiffusion \cite{wang2024instancediffusion} are state-of-the-art methods which can support controlled image generation using discrete conditions such as bounding boxes. By integrating spatial conditioning controls, these methods enable users to have control over the positioning of instances in generated images. This advancement allows for fine-grained manipulation and customization in the image generation process.
\subsection{Visual Text Generation}
Current mainstream text-to-image generation models, like Stable Diffusion, excel at producing high-quality images. However, they struggle to generate accurate and legible text on these images. To address this limitation, recent research studies \cite{ma2023glyphdraw,yang2024glyphcontrol,tuo2023anytext,cao2024controllable,chen2024textdiffuser,chu2024causal} have focused on integrating clear and readable text into images by introducing glyph conditions in the latent space. These advancements, particularly, GlyphControl \cite{yang2024glyphcontrol} and AnyText \cite{tuo2023anytext}, can be seamlessly plugged into existing diffusion models, allowing for more precise rendering of text on generated images.
\subsection{Knowledge Graph and LLM}
\textbf{Knowledge Graph (KGs)} are structured multirelational knowledge bases that typically contain a set of facts. Each fact in a KG is stored in the form of triplet $(s,r,o)$, where $s$ and $o$ represent the subject and object entities, respectively, and $r$ denotes the relation connecting the subject and object entity. KGs are crucial for various applications as they offer accurate explicit knowledge \cite{ji2021survey, wang2023enhancing,zhang2021neural,sheu2021knowledge}. \textbf{LLM}, pre-trained on the large-scale corpus, such as ChatGPT \cite{brown2020language} and GPT-4  \cite{gpt4} have showcased their remarkable capabilities in engaging in human-like communication and understanding complex queries, bringing a trend of incorporating LLMs in various fields \cite{anil2023palm,gunasekar2023textbooks,jiang2023towards,xue2023weaverbird,wang2024llmrg,chu2024llm}.
By incorporating KGs, LLMs can benefit from the extensive knowledge stored in a structured and explicit manner. This integration enables LLMs to have a better understanding of the information contained in KGs, which also enhance the performance and interpretability of LLMs in various downstream tasks \cite{pan2024unifying}. In our work, we leverage the knowledge retrieved from KGs to improve prompt analysis and object localization, enhancing the overall effectiveness of LLMs.
\section{Method}
\begin{figure*}[t]
    \centering
    \includegraphics[width=0.9\textwidth]{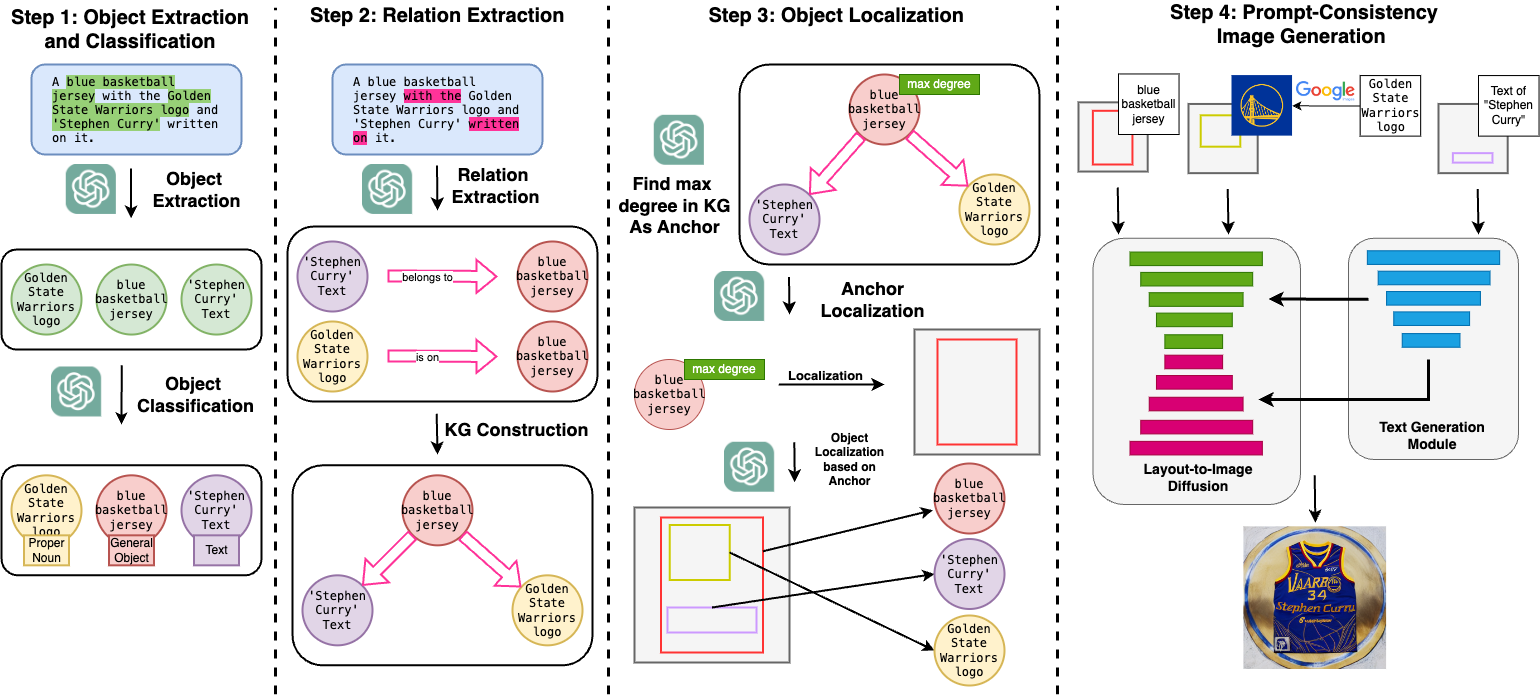}
    
    \caption{The pipeline of our PCIG method, using the example "A blue basketball jersey with the Golden State Warriors logo and 'Stephen Curry' written on it."}
    
    \label{pipeline}
    \vspace{-3mm}
\end{figure*}
In this section, we first provide a detailed definition of consistency hallucination in Sec \ref{hall_def}. Following that, we delve into the details of our framework with object extraction and classification in Sec \ref{obj_ext_cls}, relation extraction in Sec \ref{rel_ext}, object localization in Sec \ref{obj_loc}, and non-hallucinatory image generation in Sec \ref{img_gen}. More details of our PCIG framework can be seen in Figure \ref{pipeline}.
\subsection{Consistency Hallucination Definition}
\label{hall_def}
Before delving into the methodology, it is essential to first define the types of hallucinations more detailed. Based on the MHaluBench \cite{chen2024unified} benchmark, our focus is centered upon four primary types of hallucinations that arise in text-to-image generation: (1) \textbf{AH(attribute hallucinations)}, where the attributes of objects in the generated images are incongruous with the provided prompts; (2) \textbf{OH(object hallucinations)}, where the number, placement, or other aspects of objects differ from the provided prompts; (3) \textbf{SCH(scene-text hallucinations)}, where the textual content within the generated images does not align with the given prompts;(4) \textbf{FH(factual hallucinations)}, where the depicted properties of objects contradict their real-world counterparts. While there are numerous other issues related to image hallucinations, this paper focuses chiefly on the aforementioned problems.
\subsection{Object Extraction and Classification}
\label{obj_ext_cls}
\textbf{Object Extraction.} The initial step of the method involves a meticulous process of identifying objects and their attributes from the textual prompt. Given the initial prompt $P$, we extract the objects present, their quantities, and their specific properties and structured them as ${O=\{o_i\}_{i=1...N_o}}$ where $o_i$ represents a concise caption of object, comprising the combination of attribute and object name(i.e.\ A grazing sheep), and $N_o$ denotes the number of the objects identified from the initial prompt. This step is imperative since understanding the precise object information is necessary to accurately generate an image that embodies these exact details. \textbf{Object Classification.} Following the object extraction, the next is to classify each identified object in $O$ into three specific categories $C = \{GO, TEXT, PN\}$ that $O=\{o_i, C\}_{i=1...N_o}$ where $GO,TEXT,PN$ represent general objects, text within the image, and objects that refer to proper nouns existing in the real world respectively. These categories are each linked to different types of hallucinations. $GO$ is associated with attribute and object hallucinations (AH and OH), $TEXT$ corresponds to scene-text hallucinations (SCH), and $PN$ is related to factual hallucinations (FH). This categorization is critical as it not only enables us to handle each hallucination problems separately for different types of objects but also lays the groundwork for subsequent relational and spatial analyses by clearly defining the nature and context of each object within the image. 
\subsection{Relation Extraction}
\label{rel_ext}

\textbf{Relationship Recognition. }Once the objects are detected, GPT-4 determines the spatial relationships and interactions between the detected objects for the initial prompt. Let $R=\{r_i\}_{i=1...N_r}$ be the set of relationships identified from the initial prompt where $N_r$ is the number of relationships identified in the prompt. \textbf{Triple Generation. }Based on the detected objects $O$ and their relationships $R$, GPT-4 generates triples in the form of (object, predicate, object) to represent the identified relationships. The set of generated triples for the provided prompt is denoted as $T=\{t_i\}_{i=1...N_t}$, where $N_t$ is the number of triples in the initial prompt. Each triples is represented as $T=(O,R,O)$. For example, if the prompt depicts a young girl is wearing a pink dress, GPT-4 would generate a triple such as (Young Girl, is wearing, Pink Dress) where $\{Young Girl,Pink Dress\}\in O$ and $\{Is Wearing\}\in R$. \textbf{Knowledge Graph Construction. }The knowledge graph the initial prompt is then constructed using the generated triples $T$ where the objects $O$ serve as the nodes and the relationships $R$ serve as the edges. The knowledge graph for the initial prompt is denoted as $G=(V,E)$ where $V=O$ is the set of nodes (objects) and $E$ is the set of edges with $R$ being the set of all possible relationships in the initial prompt. The construction of the knowledge graph using GPT-4 is a critical step in our method. It provides a structured and detailed representation of the initial prompt, capturing the relationships and interactions between objects in a way that goes beyond simple semantic features \cite{chu2024sora}. This enriched representation proves advantageous for subsequent steps of object localization and image generation.
\subsection{Object Localization}
\label{obj_loc}
\textbf{Spatial Organization. } Building upon the relationship extraction, this part focuses on the spatial organization of objects within the canvas. The process begins by identifying the node with the max degree $V_{m}$ in the knowledge graph $G$ constructed previously, highlighting it as the pivotal object in the image composition. Determining this anchor object is critical for orienting other objects in relation to it, ensuring a coherent and realistic spatial arrangement. \textbf{Bounding Box Generation. }The meticulous spatial arrangement extends to the precise calculation of each object's placement in relation to the anchor point and throughout the canvas expanse. This phase demands a fine-tuned equilibrium to instill visual authenticity, taking into account factors like object scale and spatial positioning to construct bounding boxes that convincingly outline the locations of the entities. Let $BB=\{[x_i, y_i, w_i, h_i]\}_{i=1...N_o}$ denote the set of bounding boxes for the objects, where the tuple $(x,y)$ signifies the object's coordinates on the canvas, and $(w,h)$ indicates the object's dimensions within the space. The bounding boxes are precisely structured, conforming to exact dimensional specifications and coordinate precisions, ensuring that every object is proportionately and accurately depicted within the generated image.

\subsection{Prompt-Consistency Image Generation}
\label{img_gen}
Building upon the previously described steps, we have successfully secured a series of well-defined bounding boxes for every object on the canvas. Our objective is to leverage these bounding boxes to generate corresponding images wherein the positioning of objects closely mirrors the layout specified by $BB$. To this end, we employ a controllable text-to-image  model as our primary framework of the model that is specifically designed to accept bounding boxes as input, enabling precise manipulation of image outcomes. A visual text generation module is incorporated with the model, designated to handle linguistic elements, forming the essence of our integrated system.

Our system categorizes inputs into three segments as mentioned in Sec. \ref{obj_ext_cls}: GO, TEXT, and PN for image generation. For GO, we input both the bounding box and its caption directly into the main framework of the model. For TEXT, we input the textual content and its corresponding bounding box into a visual text generation module. This module incorporates narrative elements into the visual output. For PN, we use a search engine to find representative images of the objects. These images, along with their bounding boxes, are seamlessly integrated into the model's primary input stream. By categorizing objects and applying specific generation paradigms, our model prevents the generation of images with hallucination features. This methodical approach results in photorealistic and prompt-consistency images, eliminating the challenges posed by hallucinations. 

\section{Experiments}
\subsection{Experiments Settings}
\paragraph{Dataset.} 
\begin{figure*}[t]
    \centering
    \includegraphics[width=0.8\textwidth]{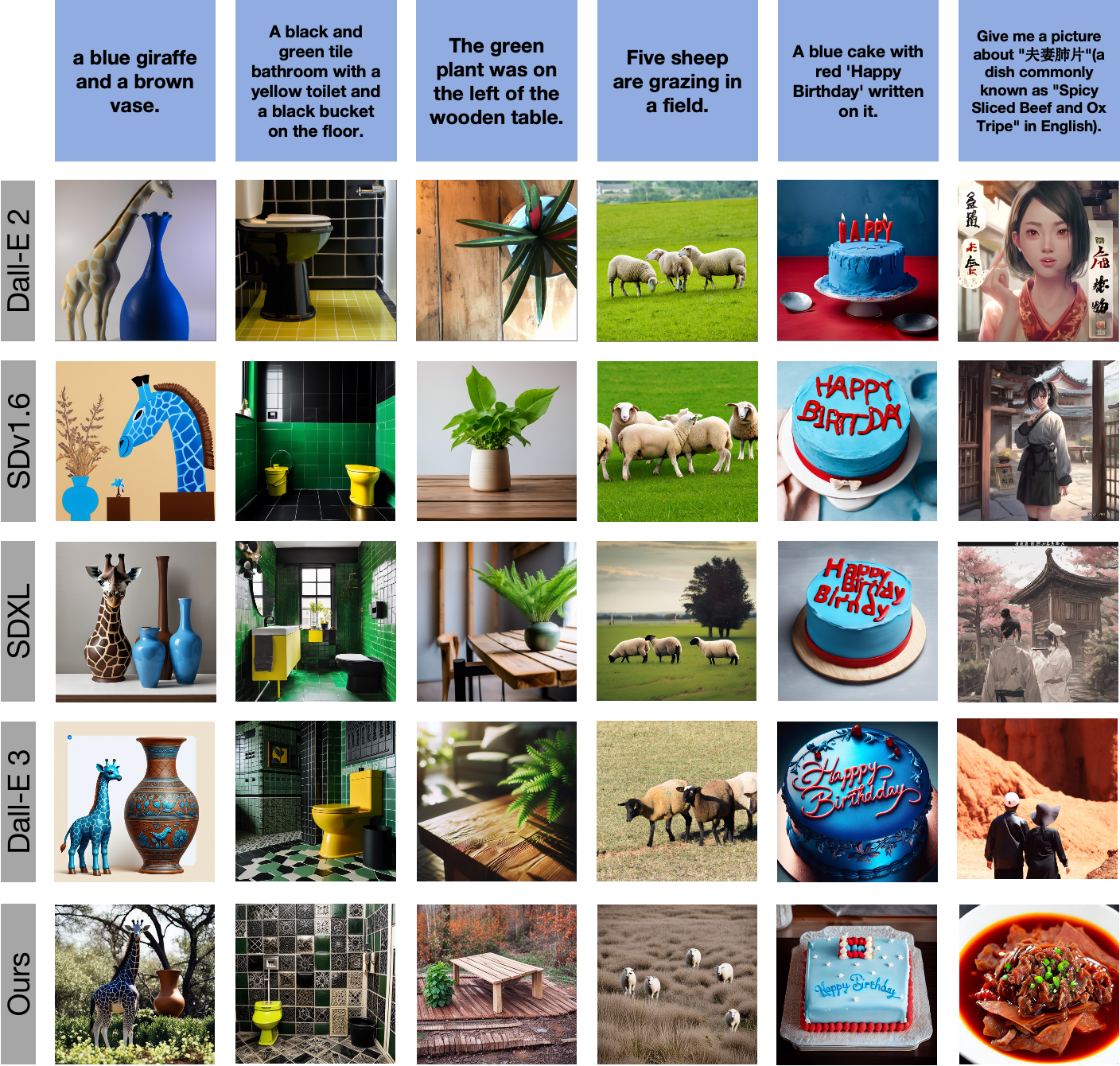}
    \vspace{-3mm}
    \caption{Compared with multiple text-to-image generation methods. Our method shows comparable performance in all aspects.}
    \vspace{-3mm}
    \label{fig_base}
\end{figure*}

\textbf{MHaluBench \cite{chen2024unified}} is a benchmark which encompasses the content from text-to-image generation, aiming to rigorously assess the advancements in multimodal hallucination detectors. The benchmark has been meticulously
curated to include 220 exemplars dedicated to Text-to-Image Generation with 158 are hallucinatory and 62 are non-hallucinatory. Specifically, it includes 137 prompts which will generate images with object and attribute hallucination potentially, 63 prompts with textual hallucination, 18 prompts with factual hallucination, 2 prompts with combination of factual hallucination and textual hallucination. In one exemplar, it contains the original prompt augmented through ChatGPT to include more specific information, the presence of hallucination, the analysis for hallucination, and the generated image based on the prompt.
\paragraph{Implement Details and Evaluation Metrics.} Our pipeline is training-free and comprises three pre-trained models. We employ the GPT-4 \cite{gpt4} as the base LLMs to generate bounding box for identified objects and choose InstanceDiffusion \cite{wang2024instancediffusion} as primary controllable text-to-image model while AnyText \cite{tuo2023anytext} as text-generation module. We utilize \textbf{UniHD \cite{chen2024unified}}, which will return a label represents whether the input image with corresponding prompt is hallucinatory or not, as our hallucination detection method for generated images. We calculate the accuracy for each hallucination type mentioned above, including object hallucination accuracy (OH acc.), textual hallucination accuracy (TH acc.), factual hallucination accuracy (FH acc.), textual and factual hallucination accuracy (TFH acc.), and overall accuracy for evaluation.
\begin{table*}[t!]
    \centering

    \resizebox{1\columnwidth}{!}{
        \begin{tabular}{lc|c |c |c|c|c}
        \toprule
         &Method
         & \multicolumn{1}{c}{OH Acc.(\%)} & \multicolumn{1}{c}{TH Acc.(\%)} & \multicolumn{1}{c}{FH Acc.(\%)}&\multicolumn{1}{c}{TFH Acc.(\%)}&\multicolumn{1}{c}{Overall Acc.(\%)} \\

        \midrule
          Text-to-Image&SDv1.6 \cite{rombach2022high} &  {15.33}   & {11.11} & {22.22}& {0.00}& {14.55} \\
          &SDXL \cite{podell2023sdxl} &    {18.98}  & {9.52}& {8.33}& {0.00}& {15.91}  \\
         &DALL-E 2 \cite{ramesh2021zero} &    {24.82}  & {7.94}& {0.00}& {0.00}& {17.73}  \\
         &DALL-E 3 \cite{DALL-E3} &    {60.58}  & {26.98}& {9.99}& {0.00}& {45.45}  \\
         \midrule
          Layout-to-Image&GLIGEN \cite{li2023gligen} &  {88.32}   & {7.94} & {22.22}& {0.00}& {59.09} \\
          &MIGC \cite{zhou2024migc} &    {94.16}  & {11.11}& {8.33}& {0.00}& {63.18}  \\
         &InstanceDiffusion \cite{wang2024instancediffusion} &    {95.62}  & {9.52}& {22.22}& {0.00}& {64.09}  \\
         \midrule
         & \textbf{PCIG (ours)} &   \textbf{94.89} & \textbf{82.54}& \textbf{77.78}& \textbf{50.00}& \textbf{89.55} \\

       \bottomrule
    \end{tabular}}
    \vspace{-3mm}
    \caption{Experimental results  of our framework and various baseline on MHaluBench dataset.}
    \vspace{-3mm}
    \label{main_results}
    
\end{table*}
\paragraph{Baseline.} Our baseline divides into two parts. The first part is the comparison with the most representative generative models, including Stable Diffusion v1.6 \cite{rombach2022high}, SDXL \cite{podell2023sdxl}, DALL-E 2 \cite{ramesh2021zero}, and DALL-E 3 \cite{DALL-E3}, which generate visually detailed images directly based on the prompt in the benchmark. The second part is the comparison with the state-of-the-art controllable text-to-image models, also named layout-to-image models, including GLIGEN \cite{li2023gligen}, MIGC \cite{zhou2024migc}, and InstanceDiffusion \cite{wang2024instancediffusion}, which introduce spatial conditioning controls to guide the image generation process. We will use the bounding box generated in the first three steps to guide the process of image generation.

\begin{figure*}[t]
    \centering
    \includegraphics[width=0.8\textwidth]{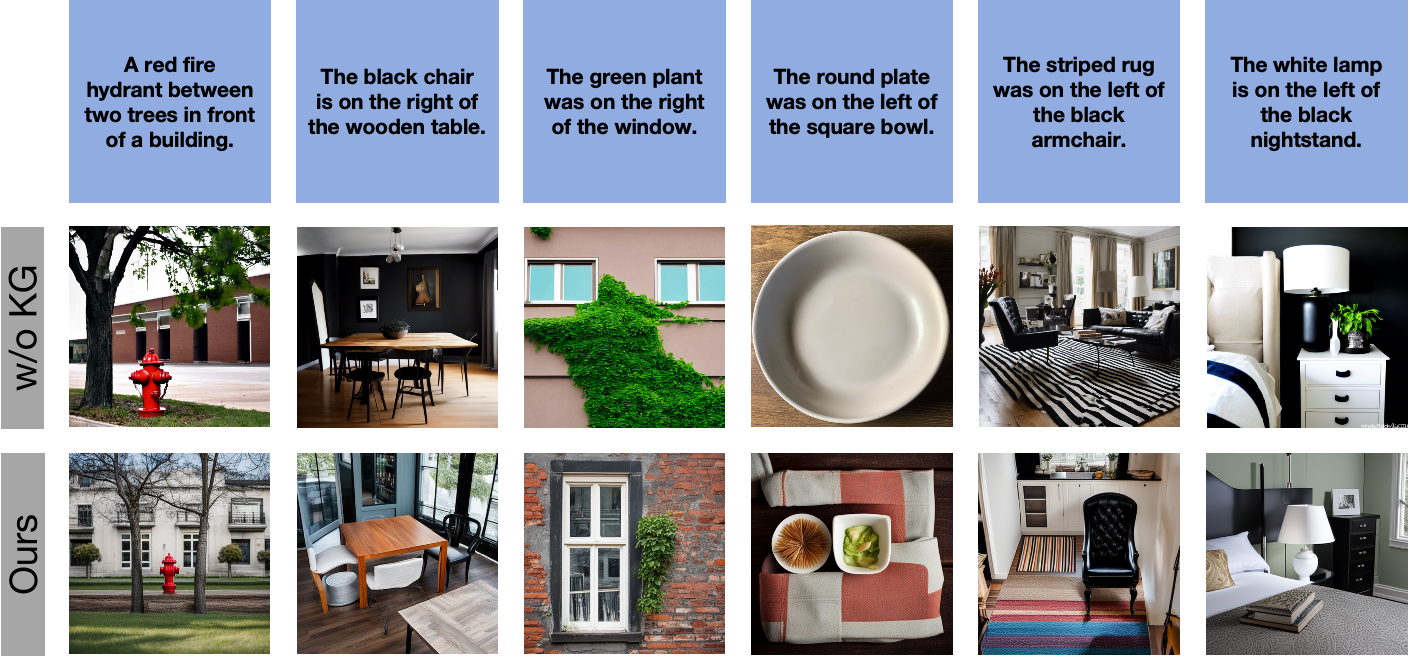}
    
    \caption{Ablation study on knowledge graph construction. Results become inaccurate in object locations when the proposed module is disable.}
    \vspace{-3mm}
    \label{w/oKG}
\end{figure*}
\begin{figure*}[t]
    \centering
    \includegraphics[width=0.8\textwidth]{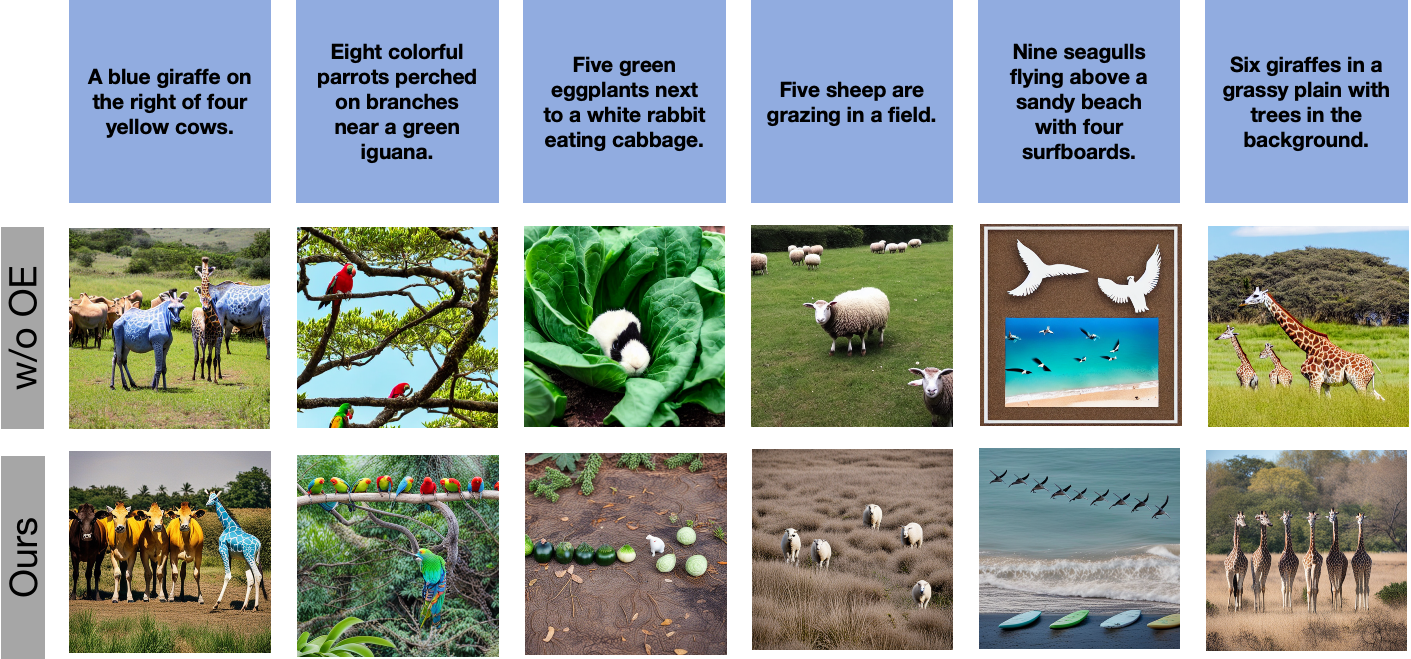}
    
    \caption{Ablation study on object extraction. Results become inaccurate in object count and attribute when the proposed module is disable.}
    \vspace{-3mm}
    \label{w/oOE}
\end{figure*}
\begin{figure*}[t]
    \centering
    \includegraphics[width=0.8\textwidth]{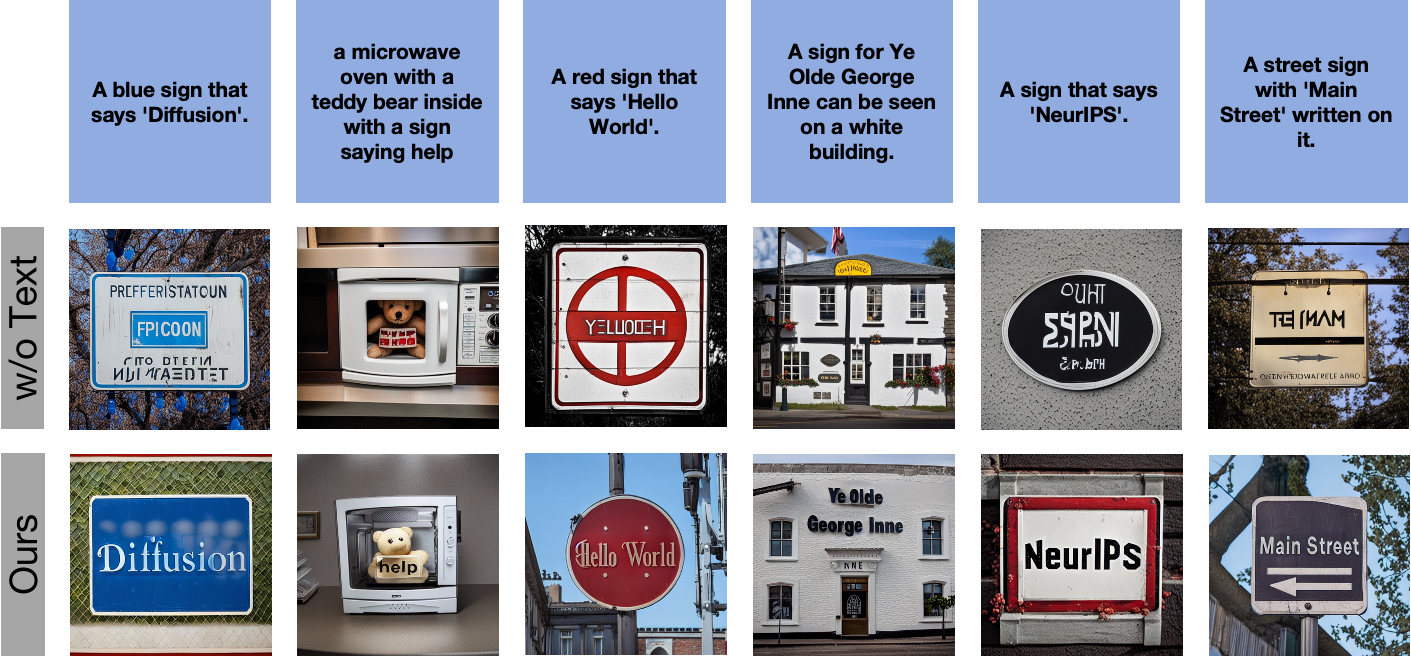}
    
    \caption{Ablation study on text generation module. Results become inaccurate in visual text when the proposed module is disable.}
    \vspace{-3mm}
    \label{w/otext}
\end{figure*}
\begin{figure*}[t]
    \centering
    \includegraphics[width=0.8\textwidth]{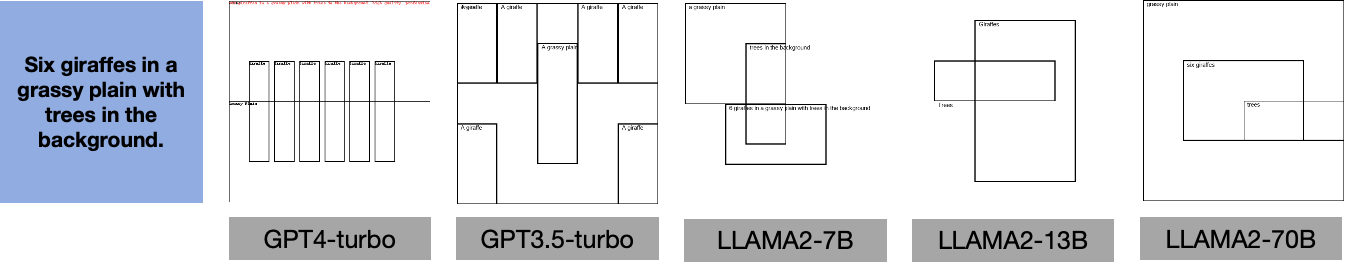}
    
    \caption{Bounding box generated by different LLM with original prompt "Six giraffes in a grassy plain with trees in the background.".}
    \vspace{-3mm}
    \label{diffllm}
\end{figure*}
\begin{table*}[t!]
    \centering

    \resizebox{1\columnwidth}{!}{
        \begin{tabular}{l|c |c |c|c|c}
        \toprule
         Method
         & \multicolumn{1}{c}{OH Acc.(\%)} & \multicolumn{1}{c}{TH Acc.(\%)} & \multicolumn{1}{c}{FH Acc.(\%)}&\multicolumn{1}{c}{TFH Acc.(\%)}&\multicolumn{1}{c}{Overall Acc.(\%)} \\

        \midrule
          w/o KG extraction &  {64.96}   & {82.54} & {77.78}& {0.00}& {70.45} \\
          w/o Object extraction &    {75.91}  & {7.94}& {11.11}& {0.00}& {50.45}  \\
         w/o Text module &    {95.62}  & {9.52}& {22.22}& {0.00}& {64.09}  \\

         \midrule
         \textbf{model (ours)} &   \textbf{94.89} & \textbf{82.54}& \textbf{77.78}& \textbf{50.00}& \textbf{89.55} \\

       \bottomrule
    \end{tabular}}
    \vspace{-3mm}
    \caption{Ablation results of our PCIG method on MHaluBench dataset.}
    \vspace{-3mm}
    \label{ablation_study}
    
\end{table*}

\subsection{Experimental Results and Qualitative Analysis}

\paragraph{Experimental Results.} Table \ref{main_results} shows that PCIG outperforms the baseline models in all metrics. It is worth noting that the object hallucination accuracy of all text-to-image models, especially in Stable Diffusion \cite{rombach2022high} and DALL-E 2 \cite{ramesh2021zero}, is extremely low. This suggests that these models struggle to generate images that align with the given prompts under such conditions. On the other hand, PCIG and other competitive layout-to-image models demonstrate exceptional abilities in accurately generating objects and their attributes in image generation. In terms of text hallucination accuracy, factual hallucination accuracy, and textual and factual hallucination accuracy, all baseline models perform poorly. In contrast, PCIG stands out from the rest. With the help of prompt analysis and text generation module, PCIG showcases exceptional performance in both text hallucination accuracy and factual hallucination accuracy. It surpasses the baseline models, highlighting its impressive capabilities in text generation and factual object generation. The corresponding prompt template is shown in Figure \ref{prompt_template}

\paragraph{Qualitative Analysis.} Through visualization of generated images, we compare our PCIG with competitive text-to-image generation models (SD \cite{rombach2022high}, SDXL \cite{podell2023sdxl}, DALL-E 2 \cite{ramesh2021zero}, and DALL-E 3 \cite{DALL-E3}). As depicted in Figure \ref{fig_base}, The first row represents the prompts given to generate images and the left column represents the model used to generate images based on prompts. As a result, Stable Diffusion, SDXL, DALL-E 2, and DALL-E 3 shows different types of visualization errors during generation, including the inconsistency between the prompts and object attributes in generated images (column 1 and 2), the inconsistency between the prompts and object locations in generated images (column 3), the inconsistency between the prompts and the number of objects in generated images (column 4), text generation error (column 5), and factual object generation error (column 6). In contrast, Leveraging the capabilities of prompt analysis and text generation module, our PCIG presents accurate and vivid images consistent with the original prompts, as shown in the last row.
\begin{table*}[t!]
    \centering

    \resizebox{1\columnwidth}{!}{
        \begin{tabular}{l|cc |cc |cc}
        \toprule
         model
         & \multicolumn{2}{c}{GLIGEN \cite{li2023gligen}} & \multicolumn{2}{c}{MIGC \cite{zhou2024migc}} &
         \multicolumn{2}{c}{InstanceDiffusion \cite{wang2024instancediffusion}}  \\
        
           \cmidrule(lr){2-3}  \cmidrule(lr){4-5} \cmidrule(lr){6-7}

         & OH Acc.(\%)  & TH Acc.(\%)& OH Acc.(\%)  
         & TH Acc.(\%)     & OH Acc.(\%)  & TH Acc.(\%) \\
       
        \midrule
          Base. & 88.32 & {7.94} & {94.16} & 11.11 & {95.62} & {9.52}   \\
           Base. w/ text module & 89.05 & {76.19} & {94.16} & 79.37 & {94.89} & {82.54}   \\
            \midrule
            \textbf{$\Delta$} & \textcolor{deepgreen}{\textbf{+0.73}} & \textcolor{deepgreen}{\textbf{+68.25}} & \textcolor{deepgreen}{\textbf{+0.00}} & \textcolor{deepgreen}{\textbf{+68.25}} & {\textcolor{red}{\textbf{-4.27}}} & \textcolor{deepgreen}{\textbf{+73.02}}   \\

       \bottomrule
    \end{tabular}}

    \caption{Comparison results of different base controllable text-to-image model with and without text module on MHaluBench dataset.}
    \vspace{-3mm}
    \label{difft2ibase}
\end{table*}
\subsection{Ablation Study}

\paragraph{w/o KG extraction.} For w/o KG extraction, the ablation experiment locate the identified objects without relation extraction and knowledge graph construction, which means the lack of relation and spatial analysis for prompts. The corresponding prompt template is shown in Figure \ref{promptwokg_template}. Table \ref{ablation_study} reveals that when the model lacks the guidance of a knowledge graph, it struggles to fully comprehend the relationship between identified objects. As a result, it is unable to provide accurate localization of objects. In Figure \ref{w/oKG}, the comparison between our method and the ablation results is depicted. The ablation experiments clearly illustrate the problems that arise when there is a lack of objects mentioned in the prompt (column 1 and 4). Additionally, they highlight the inaccuracies in the positional relationships between the objects (rest of column). These findings emphasize the importance of relation extraction and knowledge graph construction in prompt analysis.

\paragraph{w/o object extraction.} For w/o object extraction, the ablation experiment focuses on extracting relationships between objects without considering specific object information. The corresponding prompt template is shown in Figure \ref{promptwooe_template}. Table \ref{ablation_study} clearly demonstrates that when the model lacks object information, it faces challenges in accurately identifying object attributes and the number of objects while extracting relationships between them. Furthermore, it also struggles in correctly identifying object categories when generating textual and factual object. Figure \ref{w/oOE} presents a comparison between our method and the ablation results. The ablation experiment vividly highlights the problem of inconsistency between the number of objects in the generated image and the expected number of objects mentioned in the original prompt. Consequently, the model fails to provide precise object number and attribute information due to the absence of object guidance. which prove the importance of object extraction in prompt analysis.
\paragraph{w/o text module.} For w/o text module, the ablation experiment aimed to examine the impact of removing the text generation module in our model. The results, shown in Table \ref{ablation_study}, highlight that without the text generation module, the model faced challenges in generating accurate text. Figure \ref{w/otext} provides a visual comparison between our method and the ablation results. The ablation experiments demonstrate that errors, such as missing and incorrect text, were prevalent without the text generation module. These findings reinforce the significance of the text generation module in our approach.

\paragraph{Different base controllable text-to-image models.} In this section, we conducted ablation experiments using different baseline controllable text-to-image generation models, including GLIGEN \cite{li2023gligen}, MIGC \cite{zhou2024migc}, and InstanceDiffusion \cite{wang2024instancediffusion}, with and without a text generation module. Table \ref{difft2ibase} displays the outcomes of ablation experiments conducted on various controllable T2I models, focusing on object hallucination accuracy and text hallucination accuracy. The findings reveal that utilizing different models, with or without a text generation module, both yields outstanding results for object hallucination accuracy. Furthermore, the presence of a text generation module significantly enhances text hallucination accuracy. This implies that the text generation module can be seamlessly integrated into different base models to improve text generation capabilities.
\paragraph{Different LLM for prompt analysis.} In this ablation experiment, we test the performance of different language models (LLMs) for prompt analysis. The LLMs we used are GPT4-turbo \cite{gpt4}, GPT3.5-turbo \cite{brown2020language}, LLAMA2-7B \cite{touvron2023llama}, LLAMA2-13B, and LLAMA2-70B. We measure the overall accuracies of these models, and the results are summarized in Table \ref{difllmtable}. According to the results, GPT4-turbo demonstrats the highest level of competitiveness among the LLMs tested. On the other hand, LLAMA2-7B performs the least effectively compared to the other models. Figure \ref{diffllm} displays the bounding boxes generated by different language models when analyzing the prompt "Six giraffes in a grassy plain with trees in the background". GPT4-turbo accurately identifies all objects and provides reasonable positions. GPT3.5-turbo can identify all objects, but the positions it generates are unreasonable. All LLAMA model fail to recognize objects and also generate unreasonable positions.
\begin{table*}[t!]
    \centering

    \resizebox{1\columnwidth}{!}{
        \begin{tabular}{l|ccccc}
        \toprule
         Model
         & \multicolumn{1}{c}{GPT4-turbo \cite{gpt4}}
         & \multicolumn{1}{c}{LLAMA2-7B \cite{touvron2023llama}} & \multicolumn{1}{c}{LLAMA2-13B \cite{touvron2023llama}} &
         \multicolumn{1}{c}{LLAMA2-70B \cite{touvron2023llama}} &
         \multicolumn{1}{c}{GPT3.5-turbo \cite{brown2020language}} \\

        \midrule
          Overall Acc.(\%) & 89.54& 32.27 & {42.27} & {67.27} & 70.91  \\
        \midrule
          \textbf{$\Delta$} & \textcolor{deepgreen}{\textbf{+0.00}}& \textcolor{red}{\textbf{-57.27}} & \textcolor{red}{\textbf{-47.27}} & \textcolor{red}{\textbf{-22.27}} & \textcolor{red}{\textbf{-18.64}}  \\
       \bottomrule
    \end{tabular}}
    \caption{Comparison results of different LLM for our PCIG method on MHaluBench dataset.}
    \vspace{-3mm}
    \label{difllmtable}
    
\end{table*}

\section{Conclusion}
In this paper, we introduced the \textbf{P}rompt-\textbf{C}onsistency \textbf{I}mage \textbf{G}eneration(PCIG), a effective approach that significantly enhances the alignment of generated images with their corresponding descriptions. Leveraging a state-of-the-art large language module, we make a comprehensive prompt analysis and generate bounding box for each identified objects. We further integrate a state-of-the-art controllable image generation model with a visual text generation module to generate an image guided by bounding box. We demonstrate our method could handle various type of object category based on the integration of text generation module and search engine. Both qualitative and quantitative results demonstrate our superior performance.
\paragraph{Limitations.} Our method uses GPT4-turbo as our LLM to finish object extraction, relation extraction, and object localization, which costs approximately 0.08\$ in one generation process. Furthermore, our method have difficulties in generating images with complex relationship and interaction between objects as well as with small text. To address this concern, A more powerful basic diffusion model would be of great help.
\bibliographystyle{unsrt}
\bibliography{neurips_2024}


\appendix
\label{appendix}
\section{Appendix / supplemental material}
\begin{figure*}[p]
    \centering
    \includegraphics[width=1\textwidth]{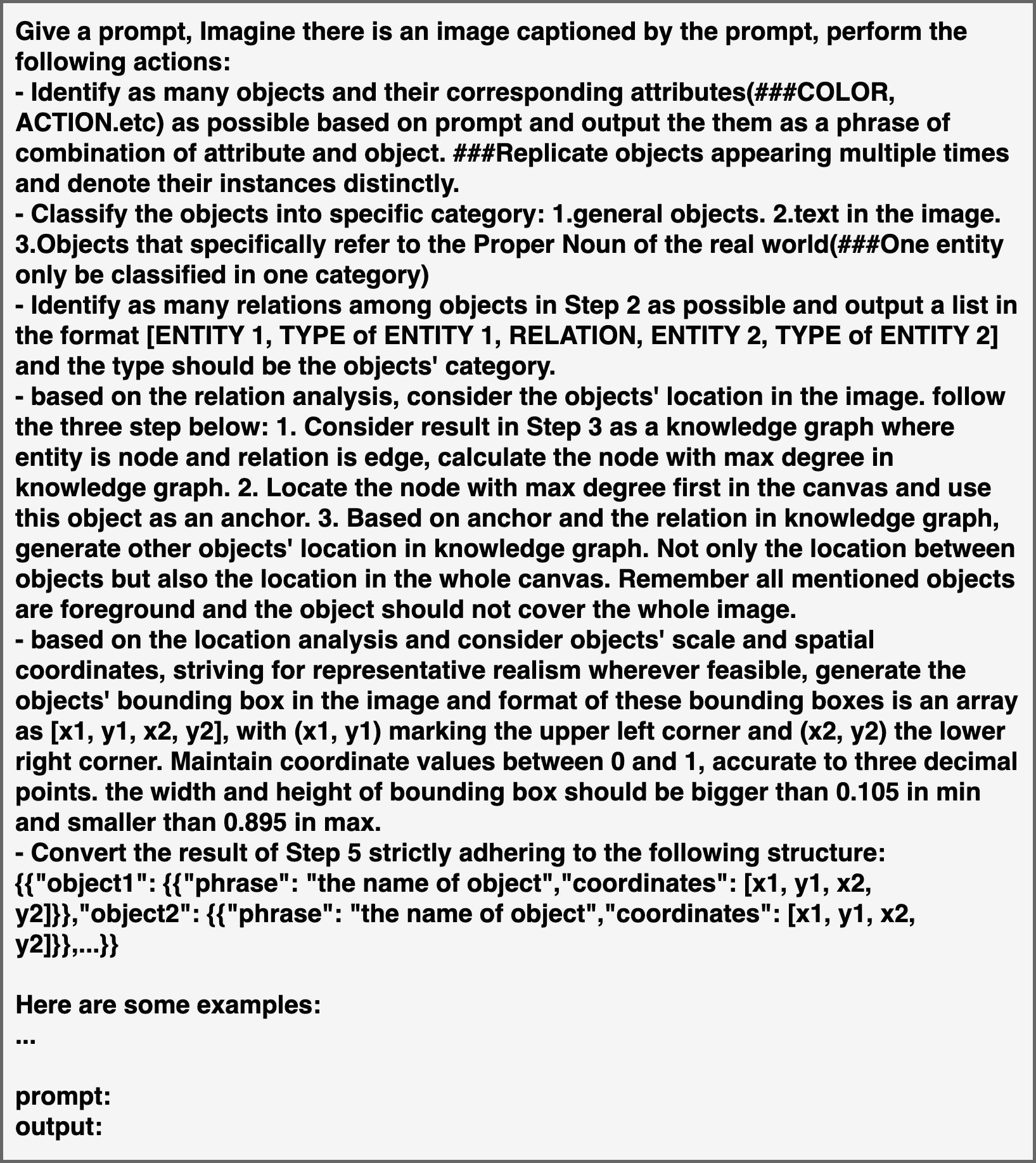}
    
    \caption{Prompt template of prompt analysis in PCIG method.}
    
    \label{prompt_template}
\end{figure*}

\begin{figure*}[p]
    \centering
    \includegraphics[width=1\textwidth]{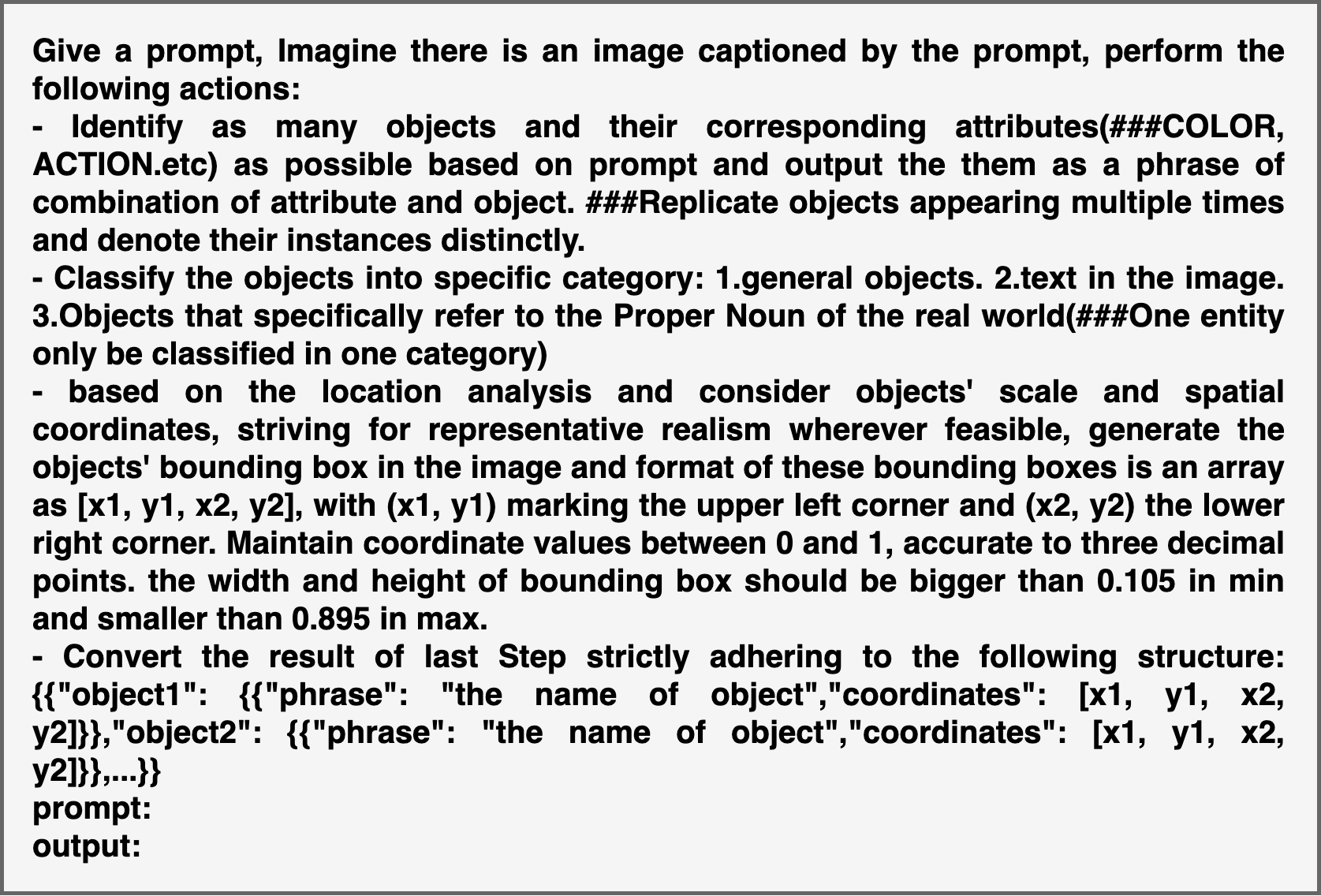}
    
    \caption{Prompt template of ablation study on knowledge graph construction in PCIG method.}
    
    \label{promptwokg_template}
\end{figure*}

\begin{figure*}[p]
    \centering
    \includegraphics[width=1\textwidth]{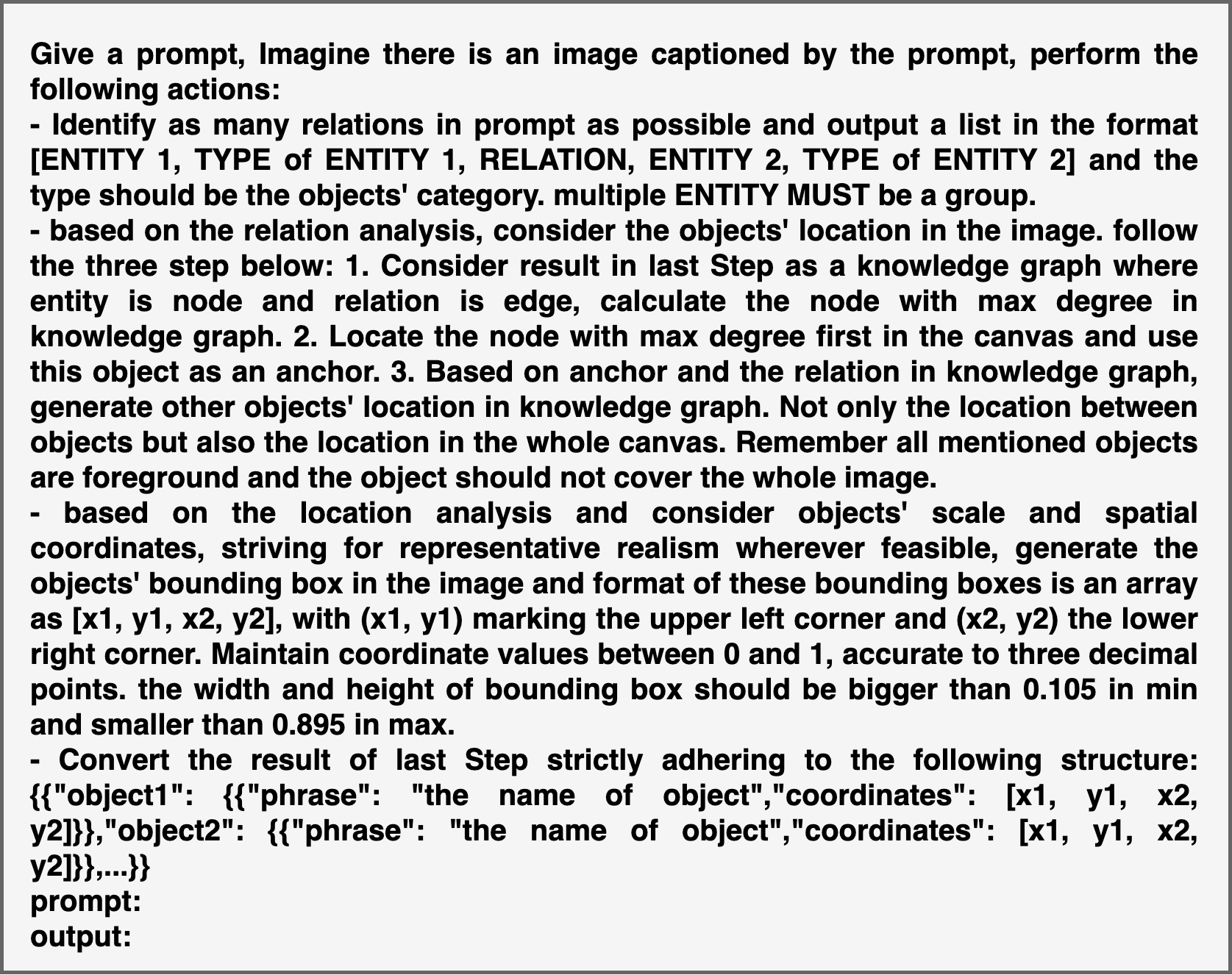}
    
    \caption{Prompt template of ablation study on object extraction in PCIG method.}
    
    \label{promptwooe_template}
\end{figure*}

\begin{figure*}[p]
    \centering
    \includegraphics[width=1\textwidth]{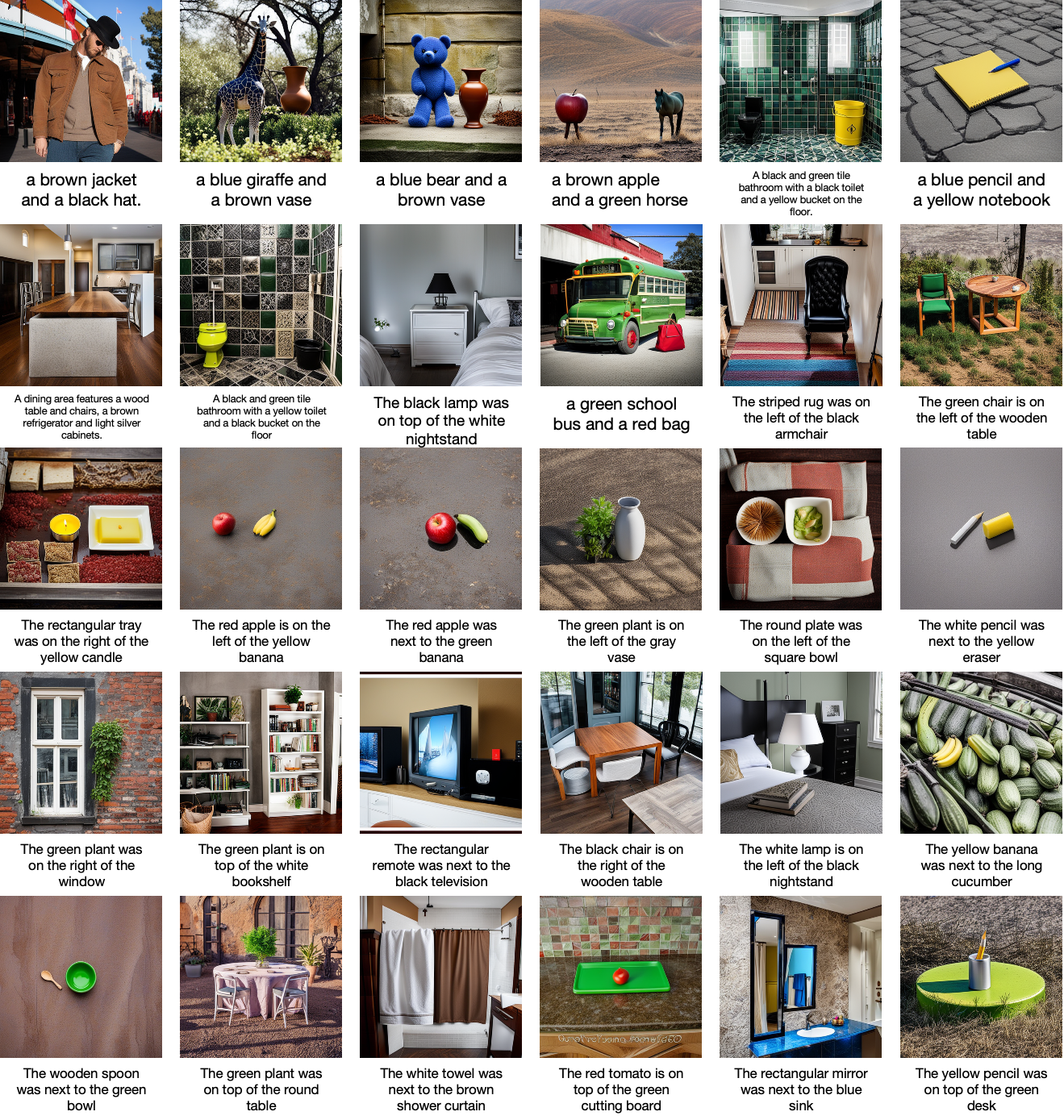}
    
    \caption{More results of our PCIG method.}
    
    \label{more_image}
\end{figure*}

In this section, we first outline the prompt template in Figure \ref{prompt_template}
designed to guide the object extraction, relation extraction, and object localization. Then we present the prompt template designed for ablation study in Figure \ref{promptwokg_template} and Figure \ref{promptwooe_template}. Furthermore, we present more results of our PCIG method in Figure \ref{more_image}.
\clearpage
\clearpage
\clearpage
\clearpage
\clearpage
\begin{figure*}[p]
    \centering
    \includegraphics[width=1\textwidth]{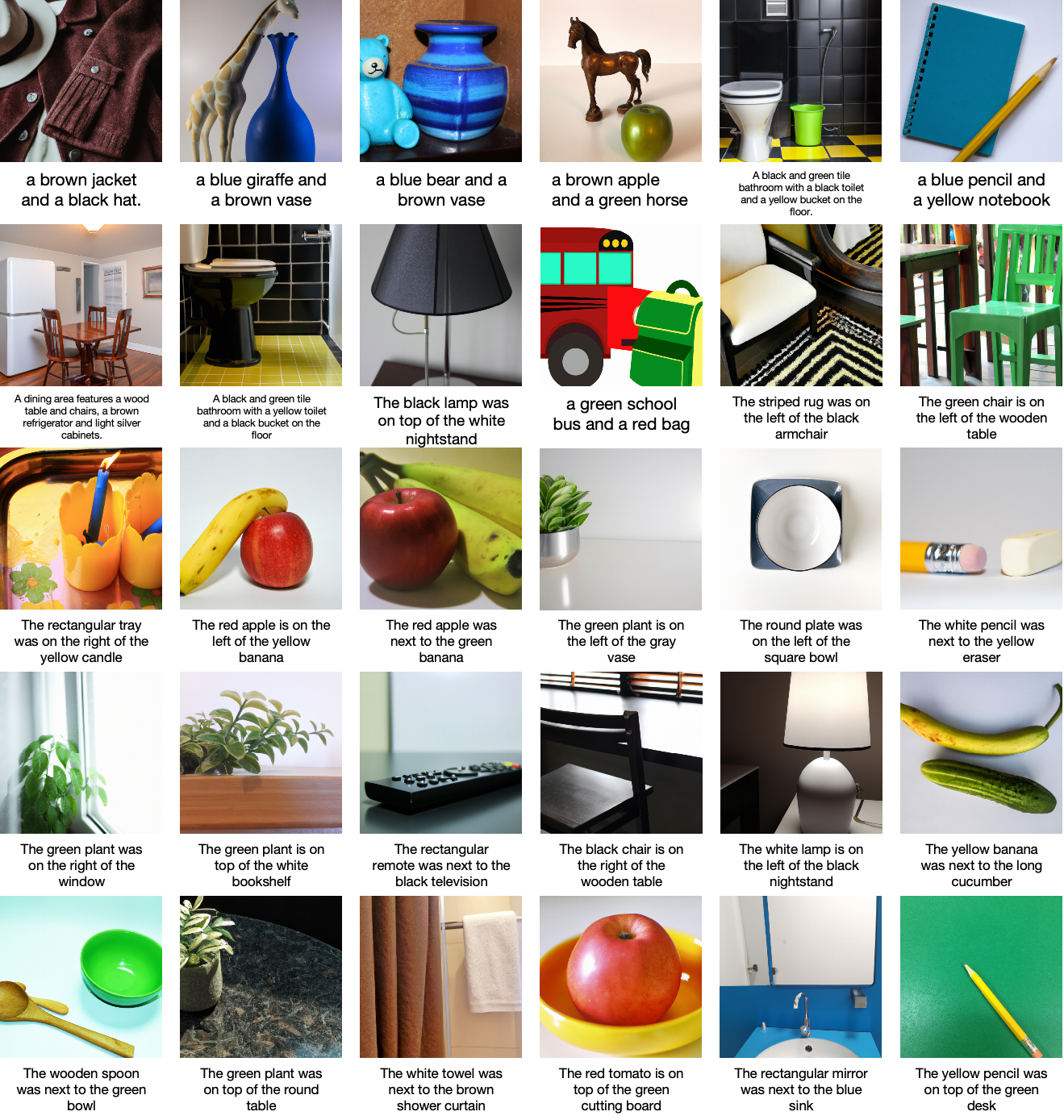}
    
    \caption{More hallucination results of DALL-E 2.}
    
    \label{dalle_more_image}
\end{figure*}
\clearpage

\newpage

\end{document}